\documentclass[sigplan,screen,9pt]{acmart}
\AtBeginDocument{%
  \providecommand\BibTeX{{%
    Bib\TeX}}}

\usepackage{algorithm}
\usepackage{algorithmicx}
\usepackage{algpseudocode}
\usepackage{multirow}
\usepackage{booktabs}

\usepackage{graphicx}
\usepackage{textcomp}
\usepackage{xcolor}
\usepackage{hyperref}

\hypersetup{hypertex=true}
\def\BibTeX{{\rm B\kern-.05em{\sc i\kern-.025em b}\kern-.08em
    T\kern-.1667em\lower.7ex\hbox{E}\kern-.125emX}}

\newcommand{\ml}[1]{{\color{red}\bf [Meng: #1]}}
\newcommand{\sz}[1]{{\color{blue}\bf [Shuzhang: #1]}}

\renewcommand\footnotetextcopyrightpermission[1]{}

\newcommand{\method}{ProPD}

\begin{document}

\title{ProPD: Dynamic Token Tree \underline{Pr}uning and Generati\underline{o}n for LLM \underline{P}arallel \underline{D}ecoding}

\author{Shuzhang Zhong}
\affiliation{%
  \institution{Peking University}
  \city{Beijing}
  \country{China}}

\author{Zebin Yang}
\affiliation{%
  \institution{Peking University}
  \city{Beijing}
  \country{China}}

\author{Meng Li*}
\affiliation{%
  \institution{Peking University}
  \city{Beijing}
  \country{China}}
\email{meng.li@pku.edu.cn}

\author{Ruihao Gong}
\affiliation{%
  \institution{Sensetime Research}
  \city{Beijing}
  \country{China}}

\author{Runsheng Wang}
\affiliation{%
  \institution{Peking University}
  \city{Beijing}
  \country{China}}

\author{Ru Huang}
\affiliation{%
  \institution{Peking University}
  \city{Beijing}
  \country{China}}

\begin{abstract}
    Recent advancements in generative large language models (LLMs) have significantly boosted the performance in natural language processing tasks.
    However, their efficiency is hampered by the inherent limitations in autoregressive token generation. 
    While parallel decoding with token tree verification, e.g., Medusa, has been proposed to improve decoding parallelism and efficiency,
    it often struggles with maintaining contextual relationships due to its independent token prediction approach and incurs significant verification overhead, especially with large tree sizes and batch processing.
    In this paper, we propose \method, an efficient LLM parallel decoding framework based on dynamic token tree pruning and generation.
    \method~features an advanced early pruning mechanism to efficiently eliminate unpromising token sequences to improve verification efficiency.
    Additionally, it introduces a dynamic token tree generation algorithm to balance the computation and parallelism of the verification phase in real-time and maximize the overall efficiency across different batch sizes, sequence lengths, and tasks, etc.
    We verify \method~across a diverse set of datasets, LLMs, and batch sizes and demonstrate \method~consistently outperforms existing decoding algorithms by 1.1-3.2 $\times$.
\end{abstract}


\maketitle

\section{Introduction}
\label{sec:intro}

Recent years have witnessed revolutionary advancements in generative large language models (LLMs)
\cite{brown2020language}, 
which can achieve start-of-the-art results in several generative natural language tasks, 
including summarization \cite{summarize}, machine translation \cite{hendy2023good}, question answering \cite{answering}, etc.
However, due to their large parameter size, complex architectures, and high computation requirements,
it is extremely challenging to deploy these LLMs in real-world applications.

\begin{figure}
    \centering
    \includegraphics[width=0.9\linewidth]{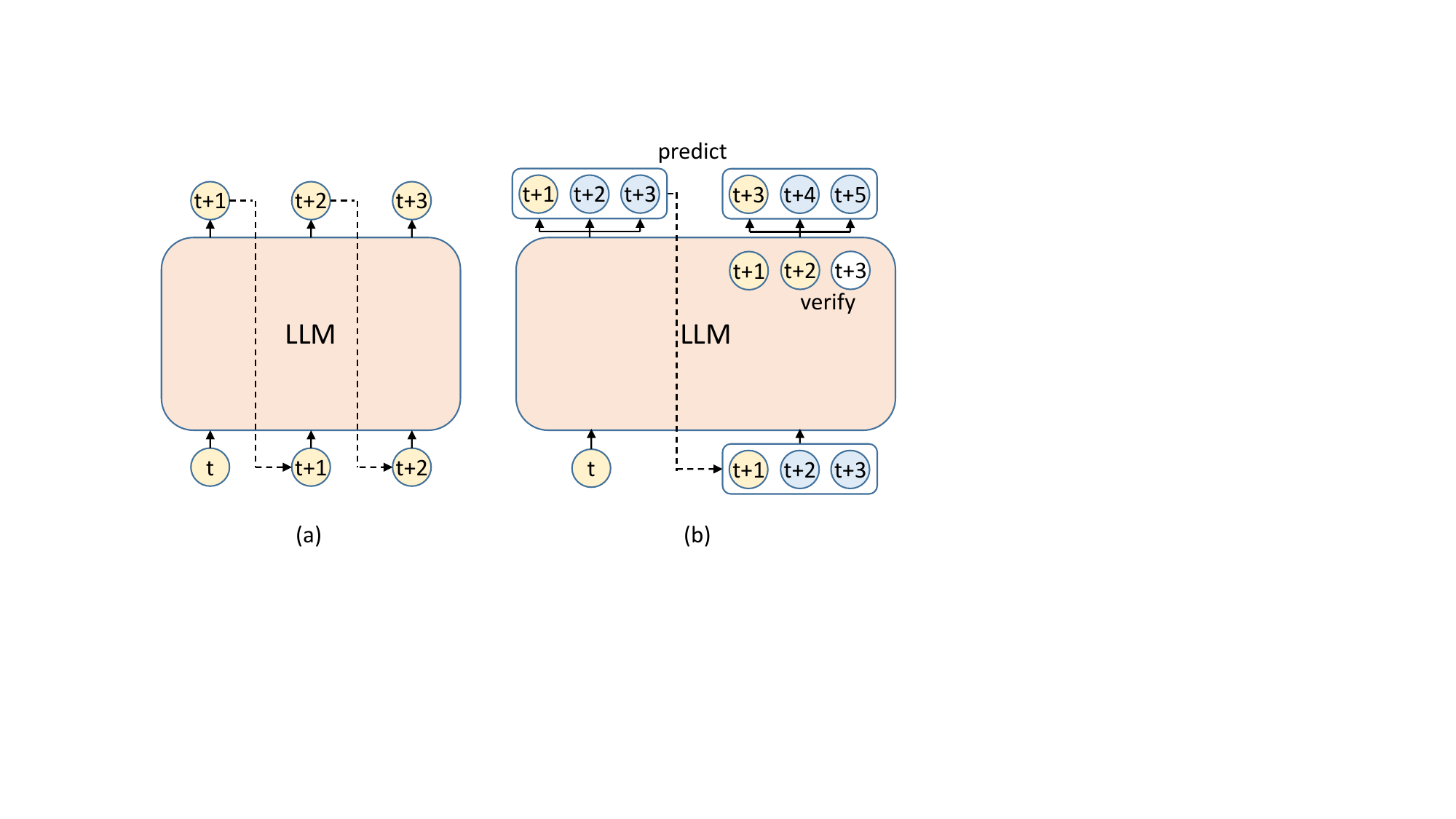}
    \caption{Workflow of (a) autoregressive decoding, (b) parallel decoding.}
    \label{fig:introduction}
    \vspace{-10pt}
\end{figure}

Modern LLMs generally leverage an autoregressive decoding algorithm \cite{radford2018improving,radford2019language,brown2020language}:
they take as input a sequence of tokens and then,
generate subsequent tokens one at a time as shown in Figure~\ref{fig:introduction} (a). 
The generation of each new token is conditioned on both the input tokens and the previously
generated tokens. While the decoding algorithm can fully capture the dependency
between tokens and preserve the context of the generated tokens,
it suffers from suboptimal runtime performance and limited GPU utilization.
This is because the degree of computation parallelism is very low and hence,
resulting in severe memory bottleneck\cite{kim2023full}.

To address the inefficiency in autoregressive token generation, parallel decoding,
e.g., Medusa \cite{cai2023medusa}, has been proposed and demonstrates a promising speedup.
Instead of decoding a single token each time,
parallel decoding first generates a sequence of token candidates, 
and then, verifies all the candidates in parallel as shown in Figure~\ref{fig:introduction} (b).
The token candidates can be further organized as a tree structure to reduce the computation
in the verification phase.
While parallel decoding increases the computation, it still achieves around $2\times$
speedup. This is because, on the one hand, LLM decoding is mainly limited by the memory bandwidth
and thus, the introduced computation incurs a small latency overhead; on the other hand,
parallel decoding can opportunistically accept more tokens at each step, and hence,
reduces the overall number of iterations.

However, we observe the existing parallel decoding method suffers from a high latency overhead
for batch decoding. Even for small batch sizes, e.g., 4, the speedup of parallel decoding
quickly diminishes. We observe the inefficiency of parallel decoding mainly comes from
two folds: \underline{1)} due to a lack of contextual relationships among the tokens generated
in parallel, a large number of tokens need to be verified, especially for large batch sizes;
\underline{2)} the generation pattern of the token candidates is static and cannot consider
the impact of batch sizes, sequence lengths, tasks, etc.

In this paper, we propose \method~to enhance LLM parallel decoding with dynamic token tree pruning
and generation. To reduce the verification overhead, we observe early LLM layers already
demonstrate good predictive capabilities that can be leveraged to prune token candidates. Hence,
we propose a dynamic token tree pruning algorithm to significantly reduce the number of candidates
without harming the number of accepted tokens. To improve the adaptability across different
decoding conditions, we propose a dynamic token tree generation algorithm that enables to 
adapt the generated token tree in real-time during decoding. Our contributions can be summarized
as follows:
\begin{itemize}
    \item We observe the inefficiency of the existing LLM parallel decoding algorithm and propose
        \method~to improve the decoding efficiency across different decoding conditions.
    \item We propose a dynamic token tree pruning algorithm to reduce the
        verification computation by more than 2$\times$ without hurting the number of
        accepted tokens.
    \item We propose a real-time algorithm to generate the token tree adaptively according to the
        decoding conditions.
    \item We verify \method~across a diverse set of datasets, LLMs, and batch sizes, and demonstrate
        \method~consistently outperforms existing decoding algorithms by $1.1$-$3.2 \times$.
\end{itemize}

\section{Preliminary}
\label{sec:prelim}

In this section, we introduce the existing parallel decoding algorithms and review the related works.
As introduced in Section~\ref{sec:intro}, parallel decoding mitigates the inefficiency of autoregressive
decoding by generating and verifying token candidates in parallel. It generally has two phases, i.e.,
the prediction phase and the verification phase.

\begin{figure}[!tb]
    \centering
    \includegraphics[width=0.9\linewidth]{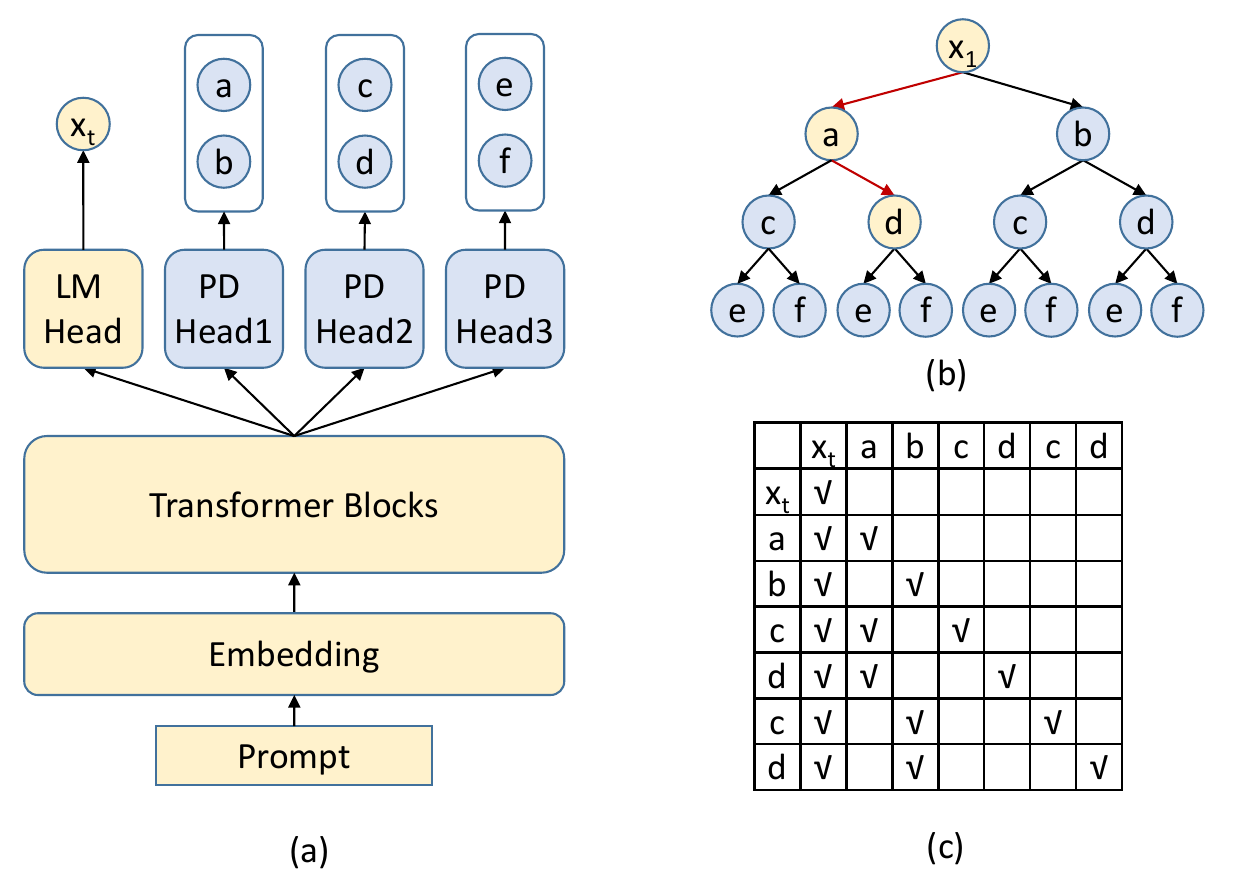}
    \caption{Example of parallel decoding: (a) model architecture, (b) token tree, (c) part of token tree mask.}
    \label{fig:parallel decoding}
\end{figure}

\paragraph{Prediction} In the Prediction phase, sequences of token candidates are predicted
with a much lower cost compared to the baseline autoregressive decoding. Depending on how
the token candidates are generated, existing works can be roughly classified into 2 categories.
The first category of works leverages a much smaller LLM to generate the candidates (sometimes
referred to as speculative decoding) \cite{chen2023accelerating, leviathan2023fast, spector2023accelerating}. While promising speedup has been demonstrated \cite{miao2023specinfer},
these works usually struggle to align the small LLM with the full-scale LLM and the requirement to
host two models in one system also drastically increases the system complexity \cite{xu2023llmcad}.
The second category leverages the full-scale LLM directly to generate the candidates \cite{stern2018blockwise,cai2023medusa,santilli2023accelerating,bae2023fast,jiang2023recyclegpt}.
A few extra model heads are trained to simultaneously predict the candidates for multiple timesteps.
While these works benefit from the system's simplicity, they also face an important limitation:
as shown in Figure~\ref{fig:parallel decoding}, the token candidates for different timesteps are generated in parallel
without considering their contextual dependency. 
This leads to an exponential increase in the generated token sequences with respect
to the number of timesteps to predict and the number of candidates at each step. For example, 
if we have parallel heads for $n$ timesteps and select the top-$k$ candidates for each timestep,
then, in total $k^n$ possible candidate sequences are generated. BPD \cite{stern2018blockwise}
select $k=1$, i.e., the most probable candidate, during generation while Medusa \cite{cai2023medusa}
leverages a heuristic method to select a different $k$ for each timestep.

\paragraph{Verification} Once a set of candidate token sequences is generated, the next step is to
leverage the full-scale LLM to verify each sequence. Due to the invocation of the full-scale LLM,
the verification phase is usually more time-consuming compared to the generation phase.
To improve the verification efficiency, existing methods\cite{cai2023medusa,miao2023specinfer} adopt token-tree verification strategies that parallelize the evaluation of multiple candidate sequences. 
Token-tree verification begins by exploiting common prefixes shared across candidate sequences as shown in Figure~\ref{fig:parallel decoding}(b), enabling the LLM to compute the initial attention and hidden states once for that prefix. Unlike traditional attention mechanisms that compute scores in a linear sequence, tree attention needs to consider the branching structure where multiple potential successor tokens may exist at the same level. To manage this, attention masks are employed to allow each token to attend only to its appropriate context as shown in Figure~\ref{fig:parallel decoding}(c).

\section{Motivation and Observation}
\label{sec:motiv}



The effectiveness of parallel decoding is concurrently influenced by both the number of accepted
token candidates (denoted as acceptance length) and the token tree verification overhead.
While a large token tree size increases the
acceptance length, it also drastically increases the verification iteration time. Hence, in order to
achieve maximum acceleration from parallel decoding, it is necessary to strike a balance between the two
impacting factors. Existing parallel decoding algorithms cannot handle the two factors well
due to a lack of contextual relationships in the sequence \cite{stern2018blockwise,cai2023medusa}.
As introduced in Section~\ref{sec:prelim}, while the candidate tokens are generated in parallel,
they have to be verified in sequences to capture the context.
Hence, directly verifying all the potential sequences results in an exponential increase in computation complexity. 
We make the following observations that motivate us to propose \method.

\begin{figure}[!tb]
    \centering
    \includegraphics[width=\linewidth]{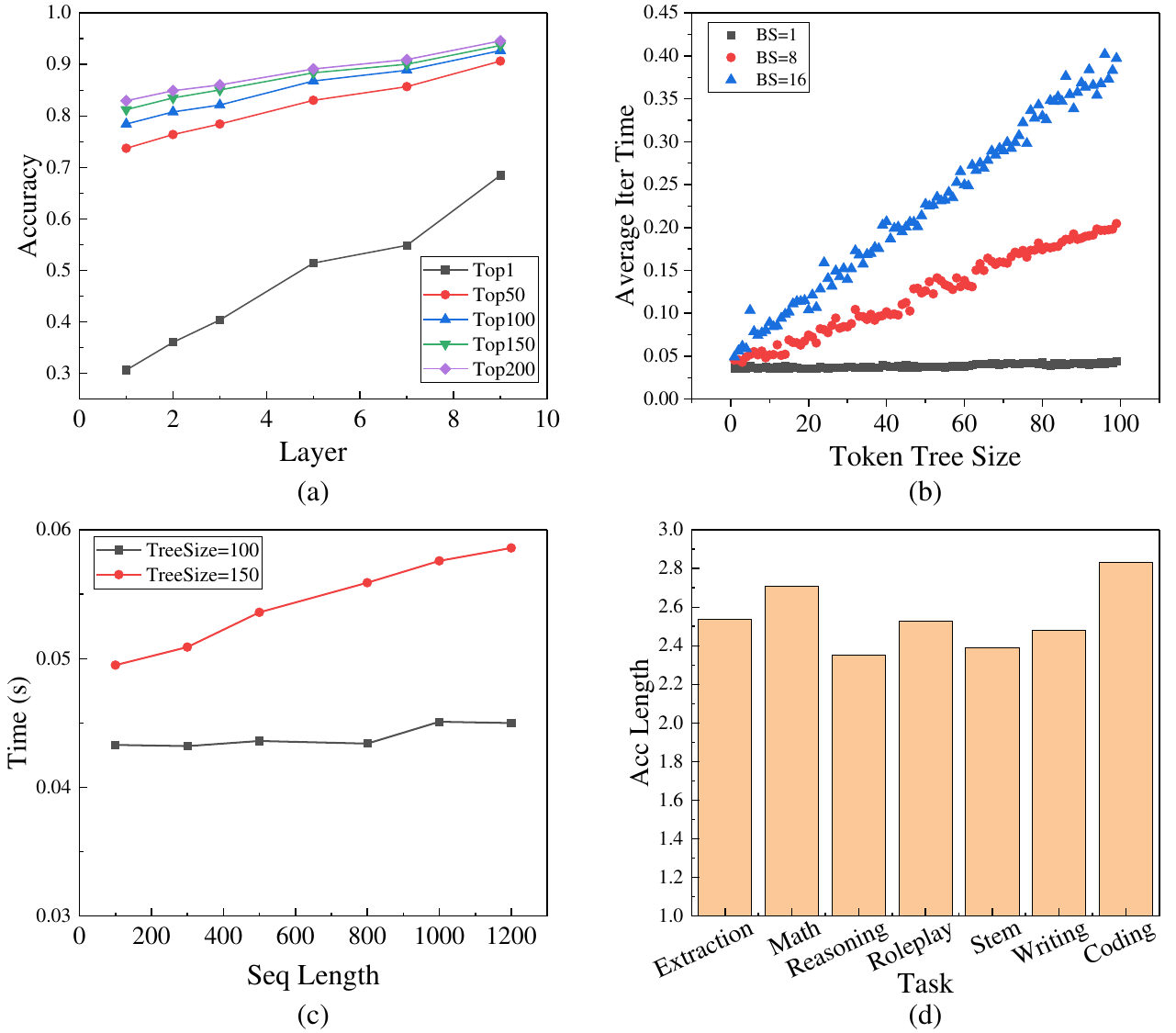}
    \caption{Iteration time and acceptance length under different scenarios: (a) Top-$k$ accuracy under different early layers, (b) average iteration time under different batch sizes and token tree sizes, (c) average iteration time under different sequence lengths, (d) average acceptance length under different tasks.}
    \label{fig:motivation}
\end{figure}

\textbf{Observation 1: early LLM layers demonstrate strong predictive capabilities.} 
Figure~\ref{fig:motivation}(a) shows the prediction accuracy of the early layers by training their own heads.
While the early layers did not exhibit high precision for their topmost predictions (Top-1), they showed a remarkable increase in accuracy within a higher Top-K range.
For example, in Layer 2, the Top-1 accuracy was approximately 37\%, but this accuracy increased substantially when considering the top 50 tokens, hinting at the layers' capacity to filter out a significant number of implausible tokens.
This trend supports the hypothesis that early layers, while not fully equipped to make the final prediction, are indeed effective in discerning a broad set of unlikely token candidates. Therefore, tokens falling outside an optimal Top-$k$ range can be pruned early, reducing the computational load in subsequent layers without significantly affecting the overall predictive accuracy.

\textbf{Observation 2: the expected speedup of parallel decoding depends on inference batch size, sequence length,
tasks, etc.} As shown in Figure~\ref{fig:motivation}(b) to (d), given a fixed token tree size, the iteration time of the verification phase
varies with different batch sizes, sequence lengths, and hardware platforms.
Meanwhile, the token accept probability also changes
for different datasets. Hence, the expected speedup of parallel decoding is directly impacted by all these factors.
BPD \cite{stern2018blockwise} proposes to only
verify the prediction with the highest probability for each token while Medusa \cite{cai2023medusa} proposes
a heuristic design for the token tree, both of which are sub-optimal due to the fixed tree size.
As shown in Figure~\ref{fig:motivation}(b), we further observe the verification iteration time scales highly linearly with the token tree size. 
This is because the computation of a transformer block, including both the fully connected layers and the attention,
increases proportionally with the token tree size.

Considering that batch size remains relatively stable and sequence length exhibits only changes gradually,
this linear scaling enables us to employ linear regression models to predict the verification time based on token tree size accurately.
This predictive capability enables us to dynamically adjust the token tree size in real time,
optimizing for inference speed and computational efficiency.

\section{\method: Parallel Decoding with Token Tree Pruning and Generation}

\begin{figure}
    \centering
    \includegraphics[width = 0.9\linewidth]{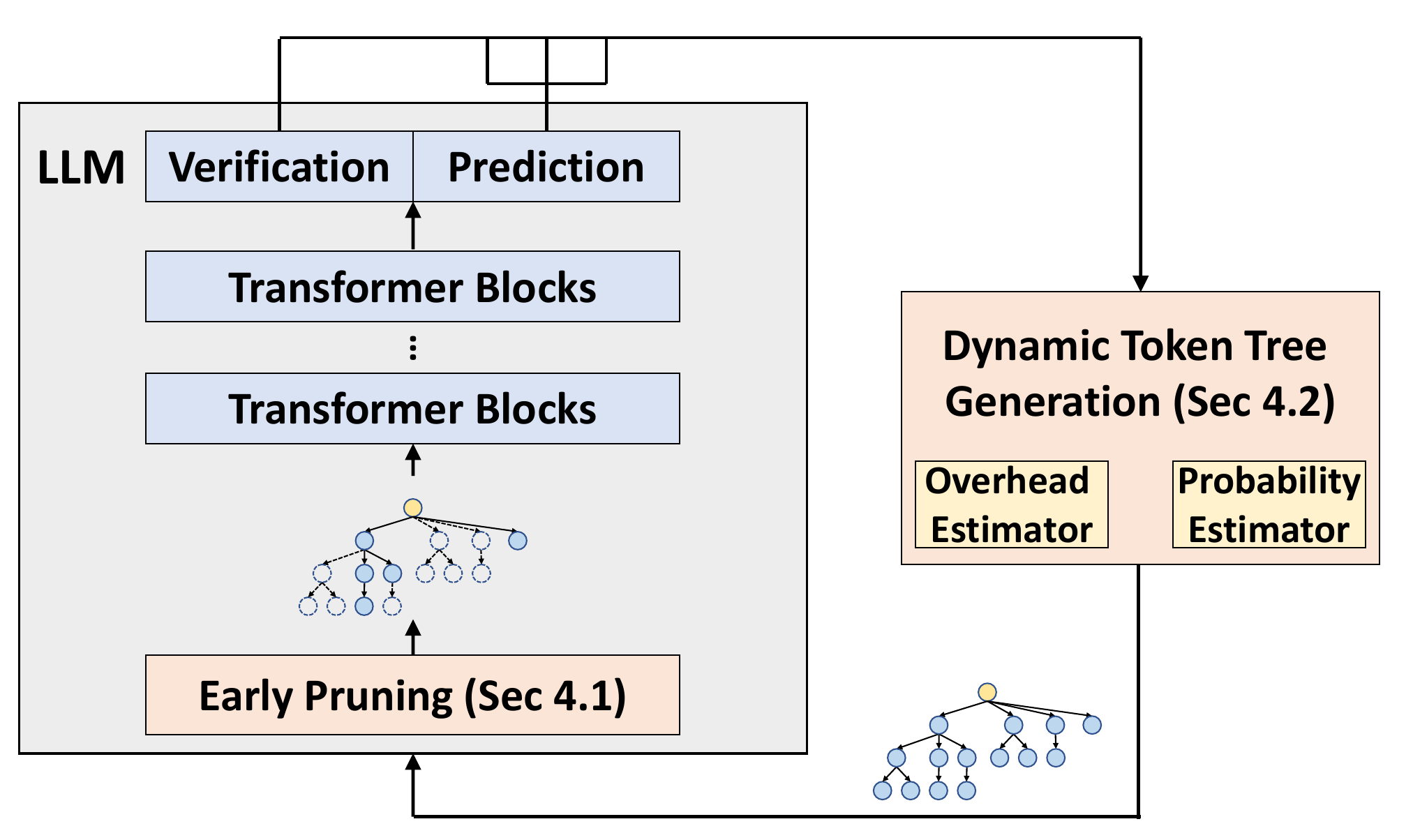}
    \caption{Overview of ProPD.}
    \label{fig:overview}
    \vspace{-10pt}
\end{figure}

The brief workflow of our framework is shown in \ref{fig:overview}.
Given the baseline parallel decoding framework, we first propose an early pruning algorithm to
remove unlikely token candidates in early LLM layers. The proposed early pruning algorithm helps address the limitations of missing contextual relationships of parallel decoding 
and reduces the computation of the verification phase.
Then, we further propose a dynamic token tree generation algorithm to enable real-time
adjustment of \method~and enables to adapt \method~to varying trade-offs in different decoding
conditions.

\subsection{Early Pruning Algorithm}

As introduced in Section~\ref{sec:prelim}, due to the lack of contextual relationships,
existing parallel decoding algorithms suffer from an exponential increase in token tree size.
While verifying the whole token tree naively incurs significant computation overhead,
Our first observation in Section~\ref{sec:motiv} indicates early LLM layers demonstrate strong
predictive capabilities, which makes it possible for the early pruning of the token tree to
reduce the computation in the verification phase. We focus on answering the following three
questions concerning the early pruning algorithm: 1) how to select the token candidates for pruning;
2) what are the key design choices to make for the pruning; 3) how to reduce the pruning
overhead to reduce decoding latency.

\begin{figure}[!tb]
    \centering
    \includegraphics[width=0.9\linewidth]{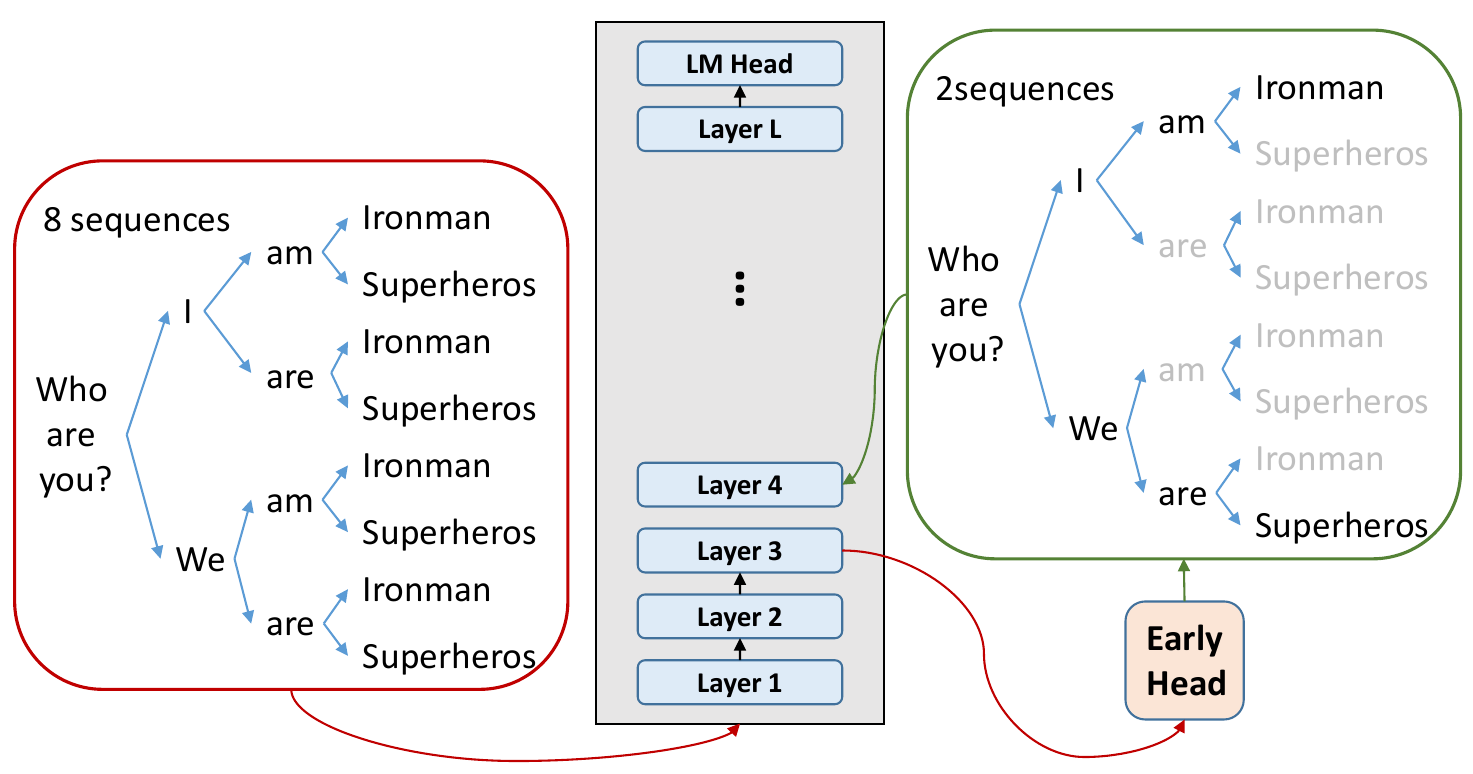}
    \caption{Early Pruning Algorithm of ProPD.}
    \label{fig:early prune}
    \vspace{-10pt}
\end{figure}

\paragraph{Pruning Criterion} To select the pruning tokens, we add an early prediction head
after a few LLM layers as shown in Figure~\ref{fig:early prune}.
We consider the following two criteria:
Top-$k$-based or probability-based selection.
Top-$k$-based selection is simple to implement and prunes all the token candidates that are not
in the Top-$k$ of the prediction directly, where $k$ is a hyper-parameter to trade-off
the pruning rate and the acceptance length. 
Probability-based selection further
leverages the predicted probability to calculate the marginal probability for each sequence
in the token three and then, either rank these sequences to prune sequences with low probabilities
or directly prune sequences with probabilities lower than a certain threshold.
We empirically find calculating the probability of each token sequence can be time-consuming since
it involves CPU and GPU communication of the predicted probability and hence, choose the Top-$k$ criterion.

\paragraph{Pruning Process} Let $n$ denote the number of LLM layers before the pruning. Then, 
the early pruning process can now be described as follows:
\begin{itemize}
    \item Prediction of successor tokens: after $n$ LLM layers, each token $x_i$
        in the token tree is processed by the early prediction head and generates a list of
        Top-$k$ most probable successor tokens, denoted as $\mathrm{TopK}(x_i)$.
    \item Token pruning: the next token $x_{i+1}$ in the sequence is evaluated against
        $\mathrm{TopK}(x_i)$ and if $x_{i+1} \notin \mathrm{TopK}(x_i)$,
        all the sequences containing $x_{i+1}$ is deemed contextually implausible.
    \item Branch elimination: collect all the tokens that fail the Top-$k$ criterion and their
        associated token sequences are pruned from the token tree.
\end{itemize}
While token tree pruning will not impact the correctness of the decoding,
the selection of $n$ and $k$ is crucial in balancing 
computational efficiency and acceptance length in our early pruning structure.
Earlier pruning layers reduce computational load but may lead to less accurate pruning decisions. 
A larger K value increases the likelihood of retaining contextually relevant sequences but also enlarges the token tree, impacting computational efficiency. 
These parameters will be empirically optimized in our experiments, aiming to find a balance that maximizes both the efficiency and accuracy of the pruning process. The experimental results will guide the final selection of these critical parameters.

\paragraph{Implementation Optimization} In the branch elimination step, token branches are eliminated
and new attention masks need to be generated. We empirically find this step can be time-consuming 
on GPU if naively implemented, e.g., re-generate the mask each time after pruning and send the mask
tensor from CPU to GPU. We propose to cache the mask on GPU and also subsample the cached mask instead
of generating a new one. Such simple optimization enables to reduce the latency overhead significantly.

\subsection{Dynamic Token Tree Generation}
As analyzed in section~\ref{sec:motiv}, the effectiveness of parallel decoding is influenced by both the acceptance length and the token tree verification overhead, which are impacted by the decoding conditions, including batch size, sequence length, etc.
Thus, we propose the dynamic token tree generation methodology to maximize decoding efficiency by balancing the length of accepted predictions and the computational overhead.
We focus on answering the following two questions concerning the dynamic generation algorithm: 1) how to estimate the computation overhead and 2) how to estimate the probable acceptance length.


\subsubsection{Verification Overhead Estimation}
Building on Observation 2 in Section~\ref{sec:motiv}, which reveals a linear relationship between the token tree size and iteration time, our framework adopts a weighted regression model for real-time estimation of this relationship.

\paragraph{Model Formulation} We formalize the weighted regression model. We denote the average iteration time for a token tree of size $i$ as $T_{perf}^i$, the estimated iteration time of size $i$ as $T_{est}^i$, the current iteration time as $t_i$, the weight function in regression model as $W_i$, the number of candidates sizes as S. The object is to estimate the regression coefficients $\beta$ to fit the $T_{est}$ as $T_{est}^i = i \beta_1 + \beta_0 \nonumber$.

\paragraph{Estimation Process} The estimation involves several steps:

\begin{itemize}
    \item Update average iteration time: given newly collected data $(i, t_i)$, the framework first updates $T_{perf}^i$ as:
        \begin{align}
        T_{perf}^i \gets (1-\alpha) T_{perf}^i + \alpha t_i \nonumber,
        \end{align}
        where $\alpha$ is a hyper-parameter to help stabilize the estimation of $T_{perf}^i$ when abnormal $t_i$ exists.
    \item Weighted regression: Next, we compute the weight for each token tree size and prioritize recent updates:
        \begin{align}
        W_i = e^{-\lambda o_i} \quad \forall i \in S \nonumber
        \end{align}
        In this equation, $o_i$ is the time since the last update of $T_{perf}^i$. Intuitively,
        $T_{perf}^i$s with more frequent updates are more important since they tend to be selected.
    \item Solve the regression model: finally, the regression model parameters can be determined as:  
        \begin{align*}
            \hat{\beta_0}, \hat{\beta_1} = \arg \min_{\beta_0,\beta_1} \sum_{i=1}^{S} W_i (T_{perf}^i - (\beta_0 + \beta_1 i))^2
        \end{align*}
        $\hat{\beta_0}, \hat{\beta_1}$ can be solved analytically with negligible latency.
\end{itemize}





\subsubsection{Probability Estimation}


To estimate the probable acceptance length with a given token tree, we track the output words from each head and record their accuracy in runtime. For each head, when a token is decoded, the framework collects how often the actual token was within the top-K predictions of that head, which is note as $P$. 
Let $\mathrm{TopK}(x_i^t)$ be the set of Top-$k$ predictions of the head $i$ at time step $t$. Once the $x_{t+i}$ is finally determined, the probability of the head i will be updated:
\begin{align*}
    P_i^k \gets (1-\alpha) \cdot P_i^k + \alpha \cdot \mathrm{1}(x_{i+1}^{t} \in \mathrm{TopK}(x_i^t)),
\end{align*}
where $\mathrm{1}(\cdot)$ is the indication function. Then we can calculate the probable accuracy of the 
$k$-th highest probability token of head $i$:
\begin{align}
    p_i^k = P_i^k - P_i^{k-1} \nonumber
\end{align}

\begin{figure}
    \centering
    \includegraphics[width=0.8\linewidth]{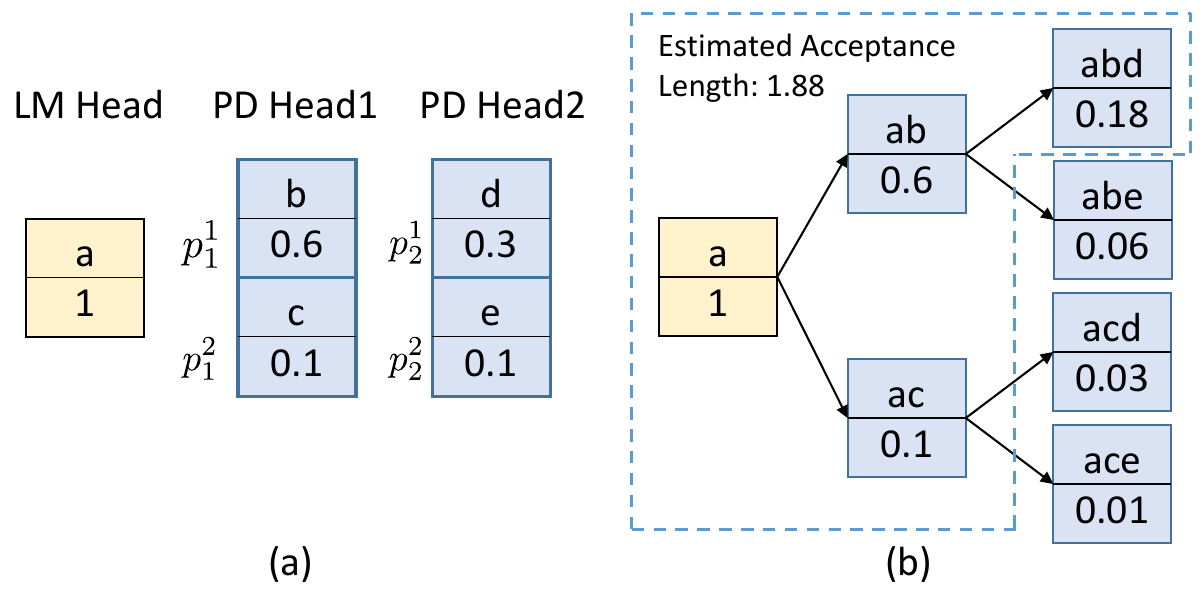}
    \caption{Probability estimation of token tree: (a) Top-$k$ probability of each head, (b) expected accuracy of each token.}
    \label{fig:probablity}
    \vspace{-5pt}
\end{figure}

Given a random sequence seq = [$s_0^{k_0},s_1^{k_1}, ... s_n^{k_n}$] generated by the parallel decoding heads, the probable acceptance length of token $s_n^{k_n}$ in sequence is:
\begin{align}
    l(\text{seq}) = \prod_{i=0}^n p_i^{k_i} \nonumber
\end{align}
in which $k_i$ means its the top $k_i$ token of head i.
For example, in Figure~\ref{fig:probablity}, the average probabilities of Top-2 tokens of the heads are shown in (a). Then, the expected acceptance length of token 'b' in sequence 'ab' is 0.6, and the expected acceptance length of token 'e' in sequence 'abe' is 0.06. If we choose 'abd' and 'ac' as the token tree, its estimated acceptance length is 1.88 as shown in Figure~\ref{fig:probablity}(b).

\subsubsection{Optimizing Efficiency}
With the verification overhead estimation and accept probability estimation, we can calculate the estimated speed of parallel decoding for each token tree size i:
\begin{align}
    v = l(i)/T_{est}^i \nonumber
\end{align}
Then we can swiftly identify the optimal token tree size for the fastest estimation by scanning the list once.
\textit{Note we do not need to invoke the dynamic token tree generation each iteration during the decoding.} Instead,
it is invoked when the batch sizes or sequence lengths change significantly. Hence, its efficiency impact is minimal and it can help
avoid expensive pre-characterization of different decoding conditions.



\section{Experimental Results}

\subsection{Experiment Setup}

\method~is implemented based on the Medusa framework \cite{cai2023medusa}.
We benchmark our framework on the open-source LLM Vicuna models \cite{mtbench}, which is finetuned on LLAMA. 
We test the 7b, 13b, and 33b models to demonstrate the scalability of \method~under different model sizes.
We evaluate our framework against the autoregressive decoding baseline, BPD, and Medusa. 
BPD \cite{stern2018blockwise} is the first parallel decoding framework.
Medusa is the SOTA parallel decoding algorithm equipped with token-tree attention.
We evaluate different frameworks on three conversational datasets: Mt Bench\cite{mtbench}, ChatGPT Prompts\cite{chatgptprompts}, and Alpaca\cite{alpaca}.
Following \cite{miao2023specinfer}, we only use the questions from these datasets to form our input prompts to simulate the real-world conversation trace.
To make a fair comparison, all of the methods take greedy decoding to guarantee the same output with the large model.
We follow \cite{cai2023medusa} and mainly evaluate the efficiency of \method~based on the number of generated tokens per second.

\subsection{Main Results}

\begin{figure*}[!tb]
    \centering
    \includegraphics[width=0.8\linewidth]{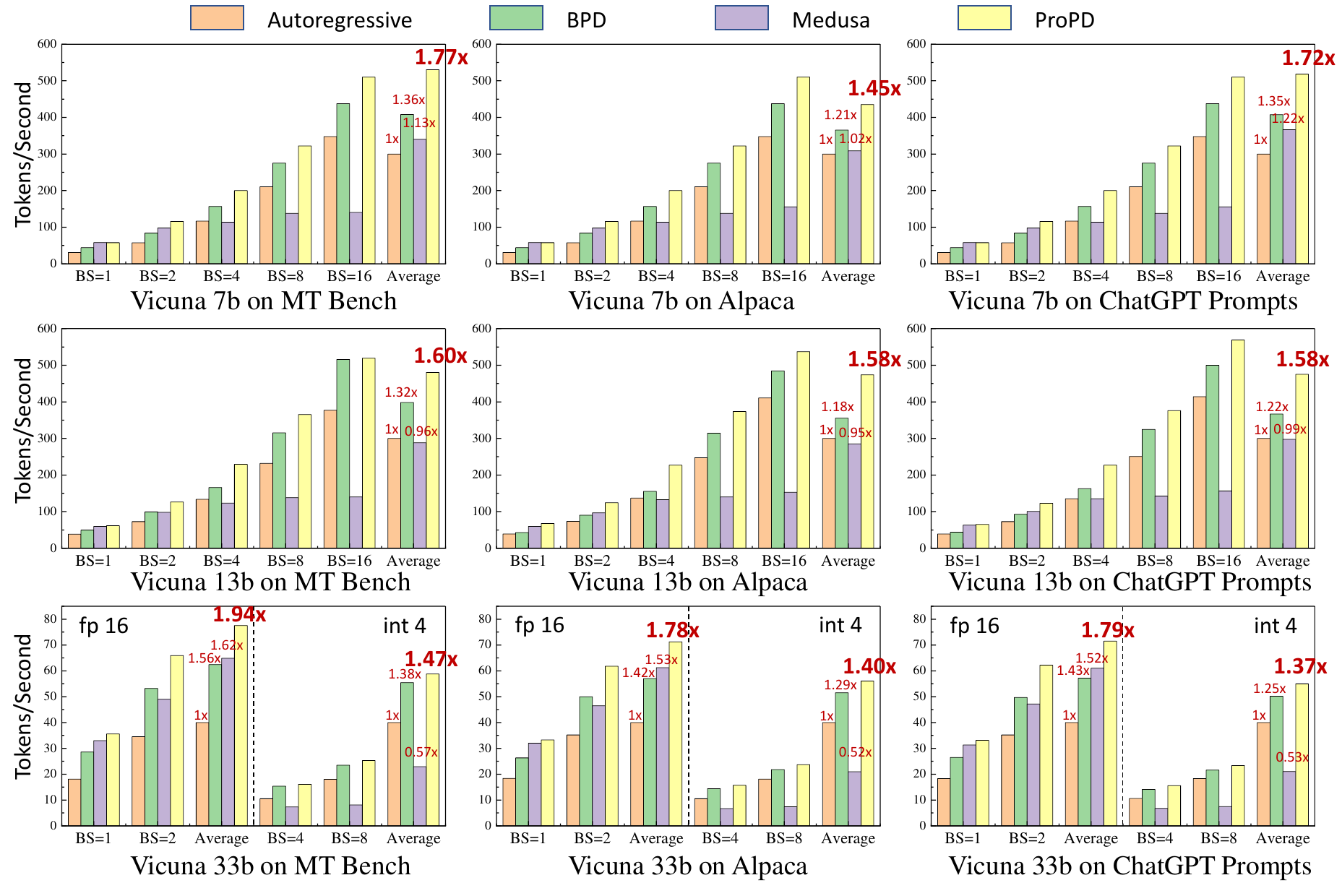}
    \caption{Inference speed comparison between ProPD and the baselines under different model sizes, tasks and batch sizes.}
    \label{fig:main result}
\end{figure*}

We investigate the generation performance of ProPD on various model sizes, datasets and batch sizes. The experiments of the 7b model are conducted on A6000, and the experiments of 13b and 33b are conducted on A100. The 33b model under batch size of 4 and 8 are loaded in 
4 bit through the transformers library to satisfy the memory limitation.    

As shown in figure~\ref{fig:main result}, the experimental results demonstrate the effectiveness of ProPD in enhancing the efficiency of parallel decoding in LLMs. The method not only accelerates the decoding process but also scales effectively with increased batch sizes, a crucial factor for practical applications. The comparison with traditional autoregressive, Medusa, and BPD methods across different model sizes and batch configurations consistently illustrates ProPD's superior performance, marking it as a significant advancement in the field of efficient language model decoding.
Table~\ref{table:speedup} shows the average speedup of ProPD against the autoregressive decoding method under different batch sizes and model sizes. ProPD can achieve 1.33-1.95$\times$ speedup under various scenarios.

\begin{table}[!tb]
\caption{Average speedup against autoregressive decoding.}
\label{table:speedup}
\centering
\resizebox{0.8\linewidth}{!}{
\begin{tabular}{c|ccccc} 
\hline \hline
           & \multicolumn{5}{c}{Batch Size}    \\ 
\hline
Model Size & 1    & 2    & 4    & 8    & 16    \\ 
\hline
7b         & 1.95$\times$ & 1.81$\times$ & 1.51$\times$ & 1.58$\times$ & 1.39$\times$  \\
13b        & 1.67$\times$ & 1.70$\times$ & 1.68$\times$ & 1.53$\times$ & 1.35$\times$  \\
33b        & 1.86$\times$ & 1.81$\times$ & 1.50$\times$ & 1.33$\times$ &   /    \\
\hline \hline
\end{tabular}
}
\vspace{-10pt}
\end{table}

\subsection{Early Pruning Accuracy}

\begin{table}[!tb]
\centering
\caption{Early pruning rate, accuracy and generation speed under BS=4 of different layers and Top-$k$ choice.}
\label{table:prune}
\resizebox{0.8\linewidth}{!}{
\begin{tabular}{ccccc} 
\hline \hline
Layer              & Top-K & Prune Rate & AccLength   &Speed\\ 
\hline
w/o pruning            &    - &  -         & 2.46        &28.43 
\\ 
\hline
\multirow{4}{*}{1}& 50   & 79.0\%    & 2.26        &43.26 
\\ 
                   & 100  & 73.0\%    & 2.32        &43.19 
\\ 
                   & 150  & 68.6\%    & 2.35        &42.98 
\\ 
                   & 200  & 64.7\%    & 2.41        &42.87 
\\ 
\hline
\multirow{4}{*}{2} & 50   & 77.3\%    & 2.32        &43.01 
\\ 
                   & 100  & 70.9\%    & 2.37        &42.94 
\\ 
                   & 150  & 65.5\%    & 2.42        &42.85 
\\ 
                   & 200  & 61.5\%    & 2.44        &42.67 
\\
\hline
\multirow{4}{*}{3} & 50   & 76.6\%    & 2.32        &42.66 
\\ 
                   & 100  & 68.3\%    & 2.44        &42.69 
\\ 
                   & 150  & 63.0\%    & 2.43        &42.48 
\\ 
                   & 200  & 59.0\%    & 2.46        &42.28 
\\ 
\hline
\multirow{4}{*}{4} & 50   & 74.0\%    & 2.43        &42.28 
\\ 
                   & 100  & 66.9\%    & 2.45        &42.20 
\\ 
                   & 150  & 62.1\%    & 2.48        &42.10 
\\
                   & 200  & 57.6\%    & 2.49        &41.94 
\\
\hline \hline
\end{tabular}
}
\end{table}


The target of the early pruning method is to maintain the original acceptance length while pruning a substantial proportion of branches. This is a critical aspect of our method's efficacy, as it balances the need for computational efficiency with the integrity of the generated token sequences.

Table \ref{table:prune} shows the pruning rate and acceptance length of ProPD under different pruning layers and topK choice. In the early layers, the acceptance length can remain close to the baseline set by Medusa with a high pruning rate. Note that the average acceptance length may increase after pruning, such as in layer 4 top 200. This is because pruning may change the sequence of positions at which the model performs parallel decoding when it prune the correct tokens. It might end up generating longer sequences at these certain positions.

We finally chose to implement early pruning at the 4-th layer of the model, with a TopK setting of 50 based on the experimental observation.

\subsection{Ablation Study}

\begin{table}[!tb]
\centering
\caption{Ablation study of ProPD.}
\label{table:ablation}
\resizebox{\linewidth}{!}{
\begin{tabular}{c|c|ccccc|c|c} 
\toprule \hline
\multirow{2}{*}{\begin{tabular}[c]{@{}l@{}}Early \\ Pruning\end{tabular}} & \multirow{2}{*}{\begin{tabular}[c]{@{}l@{}}Dynamic \\ Generation\end{tabular}} & \multicolumn{5}{c|}{7b}                                                                                                                & 13b                       & 33b                       \\ 
\cline{3-9}
                                                                          &                                                                                & \multicolumn{1}{c}{BS=1} & \multicolumn{1}{c}{BS=2} & \multicolumn{1}{c}{BS=4} & \multicolumn{1}{c}{BS=8} & \multicolumn{1}{c|}{BS=16} & BS=2                      & BS=2                      \\ 
\hline
x                                                                         & x                                                                              & \multicolumn{1}{c}{1$\times$}    & \multicolumn{1}{c}{1$\times$}    & \multicolumn{1}{c}{1$\times$}    & \multicolumn{1}{c}{1$\times$}    & \multicolumn{1}{c|}{1$\times$}     & 1$\times$& 1$\times$\\
\checkmark                                                                         & x                                                                              & 0.99$\times$& 1.18$\times$& 1.56$\times$& 1.78$\times$& 1.82$\times$& \multicolumn{1}{c|}{1.20$\times$} & \multicolumn{1}{c}{1.24$\times$}  \\
x                                                                         & \checkmark                                                                              & 1.04$\times$& 1.05$\times$& 1.50$\times$& 1.74$\times$& 2.17$\times$& \multicolumn{1}{c|}{1.17$\times$} & \multicolumn{1}{c}{1.15$\times$}  \\
\checkmark                                                                         & \checkmark                                                                              & 1.04$\times$& 1.27$\times$& 1.76$\times$& 2.34$\times$& 3.28$\times$& \multicolumn{1}{c|}{1.29$\times$} & \multicolumn{1}{c}{1.34$\times$}  \\
\hline \hline
\end{tabular}
}
\end{table}

We further conduct a breakdown analysis of the benefit brought by each of ProPD's techniques. Performance was measured across different batch sizes (BS=1 to BS=16) and model sizes (7b, 13b, and 33b). The results are illustrated in figure~\ref{table:ablation}.

\subsubsection{Early Pruning Only} 
Early pruning demonstrates excellent acceleration effects when the batch size is large. However, a slight decrease in performance was observed at BS=1 in the 7b model configuration when early pruning was applied independently. This suggests that the benefits of early pruning are less pronounced when the computational overhead is minimal. 

\subsubsection{Dynamic Generation Only}
Dynamic generation alone consistently improved performance across all batch sizes and models. This improvement underscores the efficacy of dynamic generation in enhancing the model's ability to handle multiple predictions simultaneously, thus providing a clear performance boost.

\subsubsection{Combined Early Pruning and Dynamic Generation}
The most significant performance improvements were observed when both early pruning and dynamic generation were employed simultaneously. The synergistic effect of these techniques was particularly evident at larger batch sizes (BS=4 and beyond), where the combined approach outperformed all other configurations. For instance, in the 7b model at BS=16, the performance index reached 3.28, indicating over three times the baseline performance. This because when batch size is large, the dynamic generation only method will lead to ultra small token tree and acceptance length. The pruning method can enable our framework to take larger token tree size and have bigger acceptance length.



\section{Conclusion}

In this paper, we propose \method, a framework that accelerates the parallel decoding of generative LLMs.
Existing parallel decoding methods suffer from a high latency overhead for batch decoding due to a large computation overhead.
\method~leverages a token tree early pruning algorithm to reduce the verification overhead and a dynamic tree generation algorithm
to adapt to different decoding conditions automatically.
We verify \method~across a diverse set of datasets, LLMs, and batch sizes and demonstrate $1.1 - 3.2\times$ speedup over existing
parallel decoding algorithms, e.g., Medusa.

\bibliographystyle{ACM-Reference-Format}
\bibliography{docs/reference.bib}

\end{document}